\documentclass{article}
\usepackage{ijcai24}

\usepackage{times}
\usepackage{soul}
\usepackage{url}
\usepackage[hidelinks]{hyperref}
\usepackage[utf8]{inputenc}
\usepackage[small]{caption}
\usepackage{graphicx}
\usepackage{amsmath}
\usepackage{amsthm}
\usepackage{booktabs}
\usepackage{algorithm}
\usepackage{algorithmic}
\usepackage[switch]{lineno}
\usepackage{todonotes}

\usepackage{xcolor}
\usepackage{subcaption}
\usepackage{threeparttable}
\usepackage{multirow}

\title{Layered and Staged Monte Carlo Tree Search for\\ SMT Strategy Synthesis (extended version)\thanks{This paper is an extended version of our IJCAI 2024 paper with the same title. The code and data are available at: \url{https://github.com/JohnLyu2/z3alpha}.}}

\author{
Zhengyang (John) Lu$^1$
\and
Stefan Siemer$^2$\and
Piyush Jha$^3$\and\\
Joel Day$^4$\and
Florin Manea$^2$\And
Vijay Ganesh$^3$
\affiliations
$^1$University of Waterloo\\
$^2$University of Göttingen and CIDAS\\
$^3$Georgia Institute of Technology\\
$^4$Loughborough University
\emails
john.lu2@uwaterloo.ca, stefan.siemer@cs.uni-goettingen.de, piyush.jha@gatech.edu,
\\j.day@lboro.ac.uk, florin.manea@informatik.uni-goettingen.de, vganesh@gatech.edu
}

\begin{document}

\maketitle

\begin{abstract}
Modern SMT solvers, such as \texttt{Z3}, offer user-controllable strategies, enabling users to tailor solving strategies for their unique set of instances, thus dramatically enhancing the solver performance for their use case. However, this approach of strategy customization presents a significant challenge: handcrafting an optimized strategy for a class of SMT instances remains a complex and demanding task for both solver developers and users alike.

In this paper, we address this problem of automatic SMT strategy synthesis via a novel Monte Carlo Tree Search (MCTS) based method. Our method treats strategy synthesis as a sequential decision-making process, whose search tree corresponds to the strategy space, and employs MCTS to navigate this vast search space. The key innovations that enable our method to identify effective strategies, while keeping costs low, are the ideas of layered and staged MCTS search. These novel heuristics allow for a deeper and more efficient exploration of the strategy space, enabling us to synthesize more effective strategies than the default ones in state-of-the-art (SOTA) SMT solvers. We implement our method, dubbed \texttt{Z3alpha}, as part of the \texttt{Z3} SMT solver. Through extensive evaluations across six important SMT logics, \texttt{Z3alpha} demonstrates superior performance compared to the SOTA synthesis tool \texttt{FastSMT}, the default \texttt{Z3} solver, and the \texttt{CVC5} solver on most benchmarks. Remarkably, on a challenging QF\_BV benchmark set, \texttt{Z3alpha} solves 42.7\% more instances than the default strategy in the \texttt{Z3} SMT solver.


\end{abstract}

\section{Introduction}


Satisfiability Modulo Theories (SMT) solvers~\cite{smt_introduciton_2011} are key tools in diverse fields such as software engineering~\cite{exe_2008}, verification~\cite{seahorn2015}, security~\cite{bitblaze_2008}, and artificial intelligence~\cite{vnnsmt2012}. It has long been observed that no single solver or algorithm excels across all instances of a given SMT logic or of a problem class. As a result, modern SMT solvers, such as \texttt{Z3} \cite{z3}, offer \textit{user-controllable strategies} \cite{strategy_challenge_2013}, enabling users to customize a decision procedure for their class of instances. 
A \textit{strategy} can be thought of as an algorithmic recipe for selecting, sequencing, and parameterizing \textit{tactics}. Each \textit{tactic} is a well-defined algorithmic proof rule or symbolic reasoning step, provided by the solver. 
For example, \textbf{propagate-values} is a \texttt{Z3} tactic that propagates equalities, while \textbf{sat} and \textbf{smt} are the tactic wrappers of the main SAT and SMT solver in \texttt{Z3}. A strategy builds a decision procedure by combining tactics, as shown in an exemplar strategy \textbf{(if is-pb (then propagate-values sat) smt)}. This strategy specifies a solving algorithm that, given an input instance, applies \textbf{propagate-values} followed by \textbf{sat} if the instance is a pseudo-boolean problem (as checked using \textbf{is-pb}), or applies \textbf{smt} otherwise.

Default solver strategies are typically optimized for well-established benchmarks, such as those in the SMT-LIB library \cite{smtlib}. However, as the scope of SMT applications continues to grow rapidly, users frequently encounter specialized, evolving, and unprecedented classes of instances. In these scenarios, the default or the existing customized strategies might not be as effective. Consequently, there arises a need for novel customized strategies, specifically designed to efficiently address the unique challenges posed by users' specific problems. Traditionally, this task of strategy customization has been undertaken by human experts through extensive experimentation and benchmark analysis. However, even with their expertise and efforts, the task remains challenging due to the intricate interactions among tactics and the vast search space for potential strategies. 

Early attempts have been made to synthesize SMT strategies automatically. For instance, \texttt{StratEVO} \cite{stratevo_2016} searches for an optimal strategy using evolutionary algorithms, while \texttt{FastSMT} \cite{fastsmt_2018} synthesizes a tailored strategy using imitation learning and decision tree learning techniques. While these methods show promise in automating strategy customization, they suffer from issues such as a lack of robustness, limited interpretability, and extensive training times.  

To address these issues, we introduce a novel SMT strategy synthesis method that employs Monte Carlo Tree Search (MCTS). MCTS is a heuristic search algorithm, widely applied in computer board game players as a lookahead planning algorithm \cite{mcts_survey_2012}. Its prominence further escalated following its successful integration into the groundbreaking deep reinforcement learning systems AlphaGo \cite{alphago_2016} and AlphaZero \cite{alphazero_2017}, where MCTS was employed as a policy improvement operator. 
Recently, MCTS has shown remarkable success as a standalone algorithm in solving complex symbolic or combinatorial search problems, as evidenced in Khalil et al. \shortcite{bamcts_2022} and Sun et al. \shortcite{symphylearner_2023}. Its key strengths, including its ability to effectively balance exploration and exploitation and its adaptiveness to the nuances of varied search problems, 
make it an excellent method for such challenging tasks. Our work is the first to apply MCTS to the SMT strategy synthesis problem. 

\subsection{The Strategy Synthesis Problem}
The SMT strategy synthesis problem is defined as automatically identifying an optimal strategy that yields the best performance for a given benchmark set $P$.  This performance is typically measured in terms of metrics such as the number of $P$-instances successfully solved within a specified wallclock timeout $t$. $P$ is intended to be a representative subset of the broader benchmark set $Q$, which is of interest to the user, with the expectation that a strategy performing well on $P$ generalizes effectively to $Q$. It is important to note that due to the infinite nature of the strategy search space and the empirical approach to strategy evaluation, finding a rigorously optimal solution is impractical. Consequently, our objective is to discover a near-optimal solution within a reasonable search time, based on empirical measurements. This research focuses on strategy synthesis for \texttt{Z3}, a solver that is widely regarded as one of the most prevalent SMT solvers in use today. 





\subsection{Our Contributions}

\begin{enumerate}
\item {\bf \texttt{Z3alpha}: An MCTS-based Strategy Synthesizing Solver:} We present a novel MCTS-based framework, dubbed \texttt{Z3alpha}, for the SMT strategy synthesis problem, which automatically constructs tailored solver strategies for a given class of problem instances. To the best of our knowledge, \texttt{Z3alpha} is the first MCTS-based method developed for the SMT strategy synthesis problem.


\item {\bf Layered and Staged MCTS:} To address the unique challenges inherent to strategy synthesis, that cannot be solved by the conventional MCTS alone, we develop two innovative heuristics, namely, layered search and staged search on top of the MCTS framework. The layered search method effectively narrows the search space by treating certain auxiliary tasks as independent search problems. On the other hand, the staged search technique segments the entire search problem into sequential sub-problems, enabling the use of results from early stages to expedite the search in later stages. Together, these two techniques work symbiotically to enhance the search efficiency and effectiveness in finding the optimal strategy, an essential improvement given the time-consuming nature of strategy evaluations.

\item {\bf Extensive Experimental Evaluation:} We implemented our proposed method, dubbed \texttt{Z3alpha}, on top of the leading SMT solver \texttt{Z3}. To assess its performance, we conducted comprehensive experiments, comparing \texttt{Z3alpha} with the state-of-the-art (SOTA) synthesis tool \texttt{FastSMT}, as well as \texttt{Z3}'s own handcrafted default strategy and the \texttt{CVC5} solver. 
These experiments spanned a broad spectrum, including instances from six different SMT logics (namely, QF\_\{BV, LIA, LRA, NIA, NRA, S\}), representing a wide range of problem sizes and solver runtimes. Across all experiments, \texttt{Z3alpha} consistently demonstrated superior and robust performance. This impressive performance strongly highlights the benefits of automatic strategy customization, promoting broader adoption of user-controllable strategies within the SMT community.
\end{enumerate}


\section{Related Work}
\subsection{MCTS for Symbolic/Combinatorial Problems}
MCTS has long been viewed as a powerful planning algorithm for board games \cite{mcts_survey_2012}. Recently, there has been a noticeable trend towards its application in solving symbolic and combinatorial problems. For instance, BaMCTS \cite{bamcts_2022}, an MCTS-based method, has shown remarkable success in identifying backdoors in Mixed Integer Linear Programming (MIP) problems. Symbolic Physics Learner \cite{symphylearner_2023} uses MCTS to discover nonlinear mathematical formulas for symbolic regression problems. Cameron et al. \shortcite{mcfs_2022} proposed an extension to MCTS, Monte Carlo Forest Search (MCFS), which
steers the SAT branching policy. AlphaMapleSAT \cite{alphamaplesat_2024} is an MCTS-based cube-and-conquer SAT solver. AlphaDev \cite{alphadev_2023} is an MCTS-guided deep reinforcement learning agent that synthesizes assembly programs. Remarkably, it has successfully discovered sorting algorithms that surpass the best previously known human-designed algorithms. 

Our work is, to the best of our knowledge, the first application of MCTS to address the SMT strategy synthesis problem. Different from AlphaDev, which considers the synthesis of assembly programs as sequencing individual instructions, we present the strategy program as an expression tree. The increased complexity in program structure and the extended program evaluation time present unique challenges to the strategy synthesis problem.

\subsection{MCTS Variants}
It is a well-known problem that the basic MCTS method does not generalize well between related states and actions. It results in a notably inefficient search in scenarios where the search space is extensive. To counter this issue, various techniques have been proposed. 


One prevalent technique is the rapid action value estimation (RAVE) algorithm \cite{rave_2011},
which incorporates an action-specific term $Q_{\textit{RAVE}}$ into the MCTS tree policy. Its basic assumption is that there is an intrinsic value associated with each action regardless of its position. Game abstraction \cite{gameabstraction_2013} is another common technique, especially used in Poker, to expedite the search. It abstracts similar actions or states into a single category to reduce the search space. Option Monte Carlo Tree Search (O-MCTS) \cite{omcts_2016} uses options \cite{option_1999} in the MCTS framework, to mimic the human behaviors of defining subgoals and subtasks in game playings. An option is a predefined method for reaching a specific subgoal, with its own policy and termination function. In O-MCTS, the agent selects among options instead of actions.  


Our layered and staged search methods share similarities with these MCTS variants. The layered search method reduces action and state spaces, but, instead of abstracting, it separates certain auxiliary actions aside and optimizes them in parallel. This separation generalizes values among subtrees, but, unlike RAVE which shares values between actions in different positions, it merges subtrees. The staged search chooses among previously found rewarding sub-solutions, is similar to choosing among options. However, it does not have a separate predefined policy for each choice.




\subsection{SMT Strategy Synthesis}

 \texttt{StratEVO} \cite{stratevo_2016} presents a pioneering effort in automated strategy generation, utilizing a genetic programming algorithm \cite{genetic_1994} to evolve strategies from a predefined strategy population. Unfortunately, the \texttt{StratEVO} tool is not publicly available, which precludes us from conducting an empirical comparison. 

\texttt{FastSMT} \cite{fastsmt_2018}, recognized as the SOTA tool in SMT strategy synthesis, applies a dual-phase learning approach. First, it applies the DAgger \cite{dagger_2011} algorithm to train a deep neural network (DNN), in order to discover a collection of branch-free strategies, each tailored for specific instances. These strategies are then synthesized into one single, unified strategy through an entropy-based decision-tree learning algorithm \cite{ai_2010}. Chen et al. \shortcite{synthesize4se_2021} proposed a strategy synthesis method specifically for symbolic execution, where, similar to \texttt{FastSMT}, strategies are synthesized with decision tree techniques from a list of branch-free strategies found by an offline trained DNN. 

Our method also adopts a two-step structure in our staged search approach. However, we utilize MCTS for both steps. Compared to \texttt{FastSMT}, our method shows superior and more robust empirical results across six SMT logics, as detailed in Section \ref{sec:exp}. Further, the strategies synthesized by \texttt{FastSMT} tend to be less interpretable, sometimes involving more than a thousand branches.
 
\section{Preliminaries}

\noindent {\bf SMT Solvers, SMT Logics, and SMT-LIB:}
Satisfiability Modulo Theories (SMT) solvers determine the satisfiability of first-order logic formulas, with the interpretation of symbols constrained by specific theories \cite{decisionprocedure}. SMT-LIB refers to an international initiative aimed at facilitating research and development in SMT solvers~\cite{smtlib}. The SMT-LIB initiative maintains a large library of SMT benchmarks, grouped by various SMT logics. A logic consists of one or more theories with certain restrictions, and is named after such theories and restrictions. ``QF" refers to the restriction to quantifier-free formulas, ``BV" refers to the theory of fixed-size bit-vectors, ``S" refers to the theory of strings and regular expressions, ``IA" and ``RA" refer to integer and real arithmetic. ``N" before ``IA" or ``RA" means the non-linear fragment of these arithmetics. SMT-COMP \cite{smtcomp} is an annually held competition that arose from the SMT-LIB initiative for SMT solvers. 

\noindent {\bf The Z3 Strategy Language:} The \texttt{Z3} SMT solver offers a user-controllable strategy language, allowing users to craft their customized decision procedure algorithm. A strategy selects, sequences, and parameterizes tactics, where each tactic is a built-in reasoning step in \texttt{Z3}. The context-free grammar (CFG) $G$ for the strategy language we consider in this research is shown in Figure \ref{fig:cfg_strat}, where variables are enclosed in angle brackets and terminals are highlighted in bold. 

\begin{figure}[t]
\centering
\small
\begin{align*}
\langle \text{Strategy} \rangle \rightarrow \: & \langle \text{Tactic} \rangle \\
| \: & \textbf{(using-params} \: \langle \text{Strategy} \rangle \: \langle \text{ParamSettings} \rangle \textbf{)} \\
| \: & \textbf{(then} \: \langle \text{Tactic} \rangle \: \langle \text{Strategy} \rangle \textbf{)} \\
| \: & \textbf{(or-else} \: \langle \text{Strategy} \rangle \: \langle \text{Strategy} \rangle \textbf{)} \\
| \: & \textbf{(try-for} \: \langle \text{Strategy} \rangle \: \langle \text{Constant} \rangle \textbf{)} \\
| \: & \textbf{(if} \: \langle \text{Predicate} \rangle \: \langle \text{Strategy} \rangle \: \langle \text{Strategy} \rangle\textbf{)} \\
\langle \text{Tactic} \rangle \rightarrow \: & \textbf{simplify} \:|\: \textbf{smt} \:|\: \textbf{sat} \:|\: \textbf{solve-eqs} \:|\: \textbf{elim-uncnstr} \:|\: ... \\
\langle \text{ParamSettings} \rangle \rightarrow \: & \langle \text{ParamSetting} \rangle \:|\: \langle \text{ParamSetting} \rangle \: \langle \text{ParamSettings} \rangle \\
\langle \text{ParamSetting} \rangle \rightarrow \: & \textbf{:}\: \langle \text{Parameter}\rangle \: \langle \text{Constant}\rangle \\
\langle \text{Parameter} \rangle \rightarrow \: & \textbf{som} \:|\: \textbf{flat} \:|\: \textbf{seed} \:|\: \textbf{elim\_and} \:|\: ... \\
\langle \text{Predicate} \rangle \rightarrow \: & \langle \text{BProbe} \rangle \:|\: \textbf{(}\langle \text{Operator} \rangle \: \langle \text{NProbe} \rangle \: \langle \text{Constant} \rangle \textbf{)}\\
\langle \text{BProbe} \rangle \rightarrow \: & \textbf{is-unbounded} \:|\: \textbf{is-pb} \:|\: \textbf{is-qflia} \:|\: ...  \\
\langle \text{Operator} \rangle \rightarrow \: & \mathbf{>} \:|\: \mathbf{<} \:|\: \mathbf{\geq} \:|\: \mathbf{\leq} \:|\: \mathbf{=} \:|\: \mathbf{\neq} \\
\langle \text{NProbe} \rangle \rightarrow \: & \textbf{num-consts} \:|\: \textbf{num-exprs} \:|\: \textbf{size} \:|\: ...  \\
\langle \text{Constant} \rangle \rightarrow \: & \textbf{true} \:|\: \textbf{false} \:|\: \textbf{0} \:|\: \textbf{-1} \:|\:\textbf{1} \:|\:...  \\
\end{align*}
\caption{Context-free grammar $G$ the \texttt{Z3} strategy language}
\label{fig:cfg_strat}
\end{figure}

The start symbol $\langle \text{Strategy} \rangle $ represents a strategy and is defined recursively. A strategy may consist of either a single tactic or a series of tactics linked in sequence by the tactic combinator $\textbf{then}$. Each tactic can be configured with a variety of parameters. The combinator $\textbf{or-else}$ applies the second strategy if the first strategy fails, while the combinator $\textbf{try-for}$ makes the strategy fail if it does not return within the specified timeout (millisecond). \texttt{Z3} also provides built-in probes, which evaluate formula measures, e.g., the number of constants in the formula. Predicates over them can be built using relational operators. The $\textbf{if}$ combinator constructs branching strategies based on these predicates. We refer readers to the official \texttt{Z3} guide \cite{z3guide} for more information on the strategy language. 



\noindent \textbf{Monte Carlo Tree Search:}
Monte Carlo Tree Search (MCTS) is a best-first search technique. It searches for the optimal decisions by estimating action values from numerous simulated trajectories. To search more efficiently, the method biases simulations towards previously rewarding trajectories, yet it maintains a balance by exploring less-visited paths as well. Alongside the simulations,
an MCTS tree is progressively constructed to store the action value estimations. 

\noindent Each MCTS simulation consists of 4 steps: 

\begin{enumerate}
    \item {\it Selection}: Starting from the root node, a tree-search policy traverses the MCTS tree until a leaf node is selected. The Upper Confidence Bounds applied for Trees (UCT) \cite{uct_2006} is the most common algorithm used for the tree-search policy in MCTS. UCT balances exploiting the child node with the highest value estimation and exploring the less-visited children.

\item {\it Expansion}: The MCTS tree is expanded from the selected leaf node by adding child(ren) node(s) representing unexplored actions.

\item {\it Rollout}: If the selected leaf node is non-terminal, the simulation continues by subsequently choosing actions according to a rollout policy (usually a random policy) until reaching a terminal state.

\item {\it Backup}: After evaluation, the episode reward is backed up to update the action values alongside the traversed tree path. 


\end{enumerate}



\section{\texttt{Z3alpha}: MCTS for Strategy Synthesisis}

\subsection{Modeling Strategy Synthesis as an MDP}

If we view strategy synthesis as constructing a strategy string from $G$ by sequentially applying production rules to the leftmost variable, this process can be modeled as a deterministic Markov Decision Process (MDP). An MDP is a mathematical framework in which an agent makes action decisions in a series of states, with each action leading to a new state. The agent seeks to maximize rewards over time through choices of actions. In a deterministic MDP, each action results in a deterministic state transition. 

In our formulation of the strategy synthesis problem, the {\it states} are the sentential strings derived from $G$, while the {\it actions} are the production-rule applications. The entire process of constructing a strategy is one single episode of the MDP. In an episode, the reward $R_T$ is only received at the terminal step $T$. $R_T$ is determined by the performance measure of the synthesized strategy over a given benchmark set $P$. 

Our reward system is designed to align with the evaluation criteria of SMT-COMP, prioritizing strategies that solve the highest number of instances. Simultaneously, when two strategies solve a similar number of $P$-instances, we want to steer the search towards the faster strategy. To embody these goals, we base our reward on the PAR-10 score over $P$. PAR-10 computes the average runtime for successfully solved instances and imposes a penalty for unsolved instances. The penalty is equal to the timeout value multiplied by a factor of 10. 

With this modeling, the MDP search tree is directly representative of the strategy space, and the objective becomes identifying the path with the highest reward $R_t$ in the search tree, corresponding to the optimal strategy for $P$. 


\subsection{The MCTS Framework for Strategy Synthesis}
\label{sec:framework}

We instantiate MCTS for this optimal strategy search problem. We use UCT as the tree policy in the {\it selection} phase and rollout randomly in the {\it rollout} phase. Notably, in the {\it backup} phase,  we apply the max-backup rule \cite{maxuct_2012,symphylearner_2023}. This approach updates the action values with the best observed return, rather than the average.  It encourages more aggressive exploitation towards the previously best-performing strategy, aligning with our goal. 

\begin{figure}[t]
\centering
\includegraphics[width=1\columnwidth]{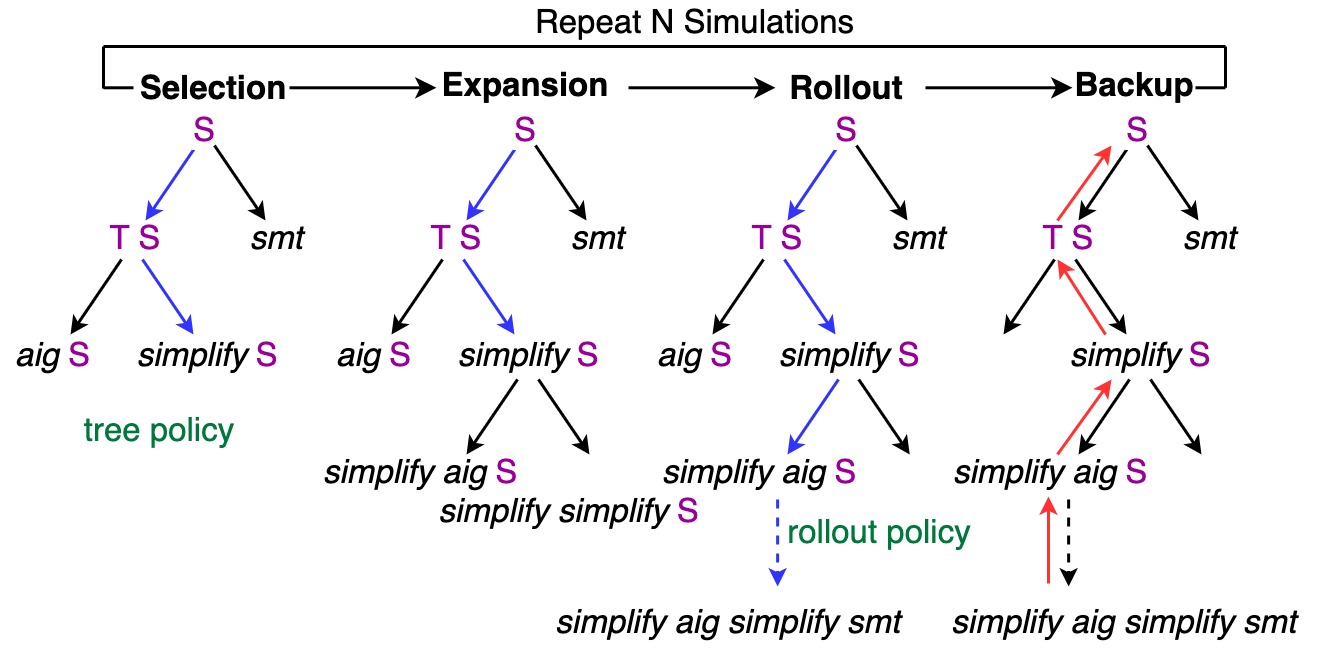}
\caption{Illustration of the MCTS framework for strategy synthesis}
\label{mcts}
\end{figure}

Therefore, the MCTS method continuously runs simulations, and in each simulation, the agent explores and assesses a single strategy, updating and retaining the best strategy seen so far. The MCTS stops when a simulation budget is reached. At the end of this process, the strategy with the highest reward $R_T$ is selected and presented as the synthesized SMT strategy for the specified instance set $P$.
Figure \ref{mcts} illustrates our basic MCTS framework, using a simplified CFG $G'$ for illustrative purposes. 
$G'$ is defined as $\text{S} \rightarrow \:  \text{T} \:\text{S}\:|\:\textbf{smt}$ and $\text{S} \rightarrow \:  \text{T} \:\textbf{simplify}\:|\:\textbf{aig}$, where $\text{S}$ and $\text{T}$ symbolize variables for strategy and tactic, respectively. 

The primary challenge in synthesizing strategies through this conventional MCTS method is the extensive time required to evaluate each strategy, which involves calling an SMT solver on all instances in $P$. This leads to a very limited exploration of potential paths, particularly given the immense search space created by the rich strategy language. To address this issue, we first add domain-knowledge rules restricting valid actions. For example, no tactic could be applied sequentially following a solver tactic such as \textbf{smt}.  
We refer readers to the Appendix for a comprehensive list of such rules. 
More importantly, we have introduced two heuristic methods, namely the layered search and the staged search, on top of the conventional MCTS, facilitating a deeper and more effective exploration of the strategy space. 

\subsection{Layered Search}
\label{sec:layered}
\begin{figure}[t]
\centering
\includegraphics[width=1\columnwidth]{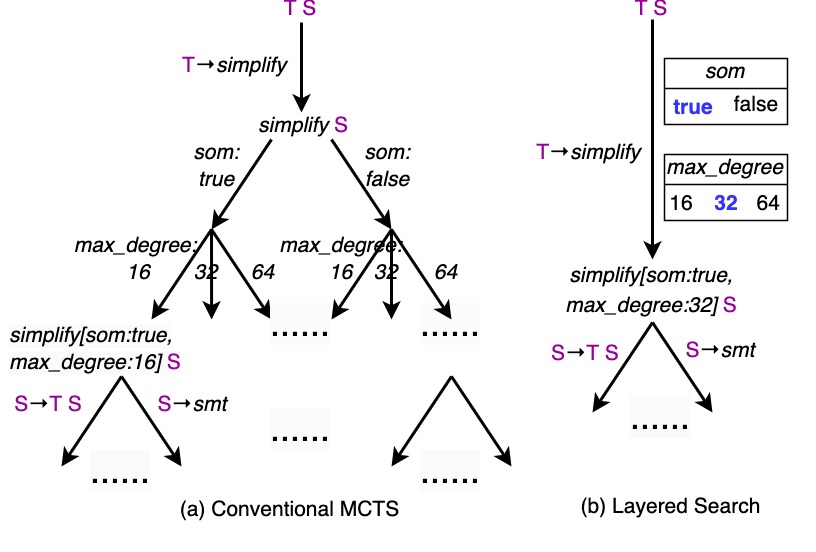}
\caption{Comparison of the conventional MCTS  and the layer-search in treating tactic parameter tuning}
\label{fig:layer_comparison}
\end{figure}

To solve the above-mentioned challenge, we propose a layered search method to optimize the tactic parameters within strategy synthesis. As shown in our CFG $G$, each tactic can be paired with multiple parameters. Using the conventional MCTS with the grammar $G$, the selection of each candidate value for a parameter is represented by one production rule, and the agent needs to make sequential production-rule decisions to configure all parameters for a given tactic, leading to exponential growth in the problem search tree, as shown in Figure \ref{fig:layer_comparison}(a).

To address this issue, our layered search method approaches the tuning of each tactic parameter as a separate Multi-Armed Bandit (MAB) problem \cite{mab1952}.  As shown in Figure \ref{fig:layer_comparison}(b), the two parameters \textbf{som} and \textbf{max\_degree} for one application of the tactic \textbf{simplify} are modeled as two MABs respectively. Each arm in the MAB represents one candidate value. For example in Figure \ref{fig:layer_comparison}(b), arm 16, 32, and 64 in the MAB \textbf{max\_degree} are three pre-selected candidate values for this parameter. 
Note that the parameter MABs are associated with a tree edge, corresponding to one specific application of a tactic, not to this tactic in general. For instance, \textbf{simplify} may be applied serval times in the strategy building. For each application, there will be two MABs representing the tuning of \textbf{som} and \textbf{max\_degree} associated with it.

One key point is that these parameter-tuning MABs are not part of the main MCTS tree.
They are engaged to select parameter values when their associated tree edge is traversed, and they are updated based on the episode reward during the {\it Backup} phase. However, such MABs do not expand the MCTS search tree after the parameter configuration, since they are separate components from the main search tree. This is in contrast to conventional MCTS, which also employs MAB principles to select among children nodes to explore, where these nodes constitute part of the search tree.  
For example, in Figure \ref{fig:layer_comparison}(a), the search tree is expanded sixfold to accommodate all possible combinations of these two parameters in the conventional MCTS framework. In contrast, in the layered search framework (Figure \ref{fig:layer_comparison}(b)), MABs for the two parameters are isolated from the search tree, creating no additional branches in the tree. 

The rationale behind the layer search is twofold. 
Firstly, tactics such as \textbf{simplify} may have dozens of parameters, and it is common for a tactic to be used multiple times within a strategy. Thus, navigating a search space that is fully expanded by all possible parameter combinations becomes impractical, especially given the time-intensive nature of strategy evaluation. 
Secondly, we argue that parameter tuning, although important, serves more as an auxiliary task in comparison to the tasks of tactic selection and sequencing. By employing the layered search method, we maintain the primary focus on the more important task. At the same time, the isolated MABs efficiently optimize the parameters without overwhelming the main search process.




\begin{table*}[ht]
\centering
\fontsize{9pt}{11pt}\selectfont
\setlength{\tabcolsep}{5pt}
\begin{tabular}{cllccccccc}
\hline
\textbf{Timeout(s)} & \multicolumn{1}{c}{\textbf{Logic}} & \multicolumn{1}{c}{\textbf{Benchmark}}  & \textbf{Test Size} & \textbf{\texttt{Z3alpha}} & \textbf{\texttt{Z3alpha0}} & \textbf{\texttt{FastSMT}}  & \textbf{\texttt{Z3}} & \textbf{\texttt{CVC5}}& \textbf{\texttt{Z3str4}}\\ \hline
\multirow{8}{*}{10}  & \multirow{2}{*}{QF\_BV}   & \texttt{Sage2} & 6444& \textbf{52.9} & 41.3 & \underline{51.3}     & 37.1     & 36.6 & -  \\ 
  &   & \texttt{core} &270 & \underline{99.6}&\underline{99.6}& \textbf{100.0}&  75.6& 82.6 & -\\ \cline{2-10} 
& \multirow{2}{*}{QF\_NIA}& \texttt{AProVE} & 1712& \textbf{94.8} &\underline{92.5}& 90.3  &  90.0 & 69.9 &-  \\ 
   &  & \texttt{leipzig} & 68 & \textbf{91.2}  & 89.1 & 88.2   & \underline{89.7} & 25.0 &-\\ \cline{2-10} 
 & QF\_NRA & \texttt{hycomp} & 1982& \textbf{91.4}  & \underline{90.8}  & 89.1 & 84.7 & 85.5 &-  \\ \cline{2-10} 
 & QF\_LIA & entire logic &12476 & \textbf{76.2}  & -  & 31.2     & \underline{74.6} & 64.8  &- \\ \cline{2-10} 
 & QF\_LRA  & entire logic&1003 & \underline{73.6}  & -  & \textbf{74.0}    & 71.0    & 63.6  &- \\ \cline{2-10} 
 & QF\_S & entire logic&18173& \textbf{99.0}  & -  & -  & 98.2 & \underline{98.4} & 97.2  \\ \hline
60    & QF\_BV  & \texttt{Sage2} & 6444& \underline{69.8}  & - & 67.7 & 57.7 & \textbf{77.8} &-\\ \hline
300   & QF\_BV   & \texttt{Sage2} & 6444& \underline{75.8}  & - & 72.6  &  69.8 & \textbf{93.3} &-       
\\ \hline              
\end{tabular}
\caption{{\bf \texttt{Z3alpha} vs. SOTA Solvers:} Percentage (\%) of instances solved from the selected SMT-LIB benchmarks across six SMT logics (
In each experiment, the result of the leading tool is highlighted in bold, while the second-best is underscored.)
}
\label{table:solved}
\end{table*}

\subsection{Staged Search}
\label{sec:staged}

While the layered search method successfully narrows down the search space, it does not alleviate the issue of significant time consumption required for each simulation evaluation. To discover a complex strategy, especially one involving nested branching, MCTS must expand over a broad space and delve deeply, often resulting in an impractically lengthy search time. This is where the staged search comes into play. 

The staged search method divides the entire search process into two stages. In the first stage, the MCTS focuses on finding high-performing {\it linear strategies} (actions introducing branches are not considered in the first stage). Here, a linear strategy is defined as a sequence of tactics without branching, where tactics are only connected with the combinator $\textbf{then}$. In the second stage, the MCTS looks for a single best strategy that combines these selected linear strategies. In other words, the second-stage MCTS works with the full grammar of $G$ but restricts the actions so that every explored strategy is a combination of the selected linear strategies through branching. By doing so, the key advantage is that the evaluation of the combined strategies can be done speedily based on the cached linear-strategy performances in the first stage, without costly SMT solver calls.


This is possible because, when executing any branched strategy $S_c$ on a given instance $f$, there is always an equivalent sequence of branch-free strategy applications, $[S_0, S_1, ..., S_N]$. For example, if we apply \textbf{((if is-pb (or-else (try-for (then simplify sat) 4000) smt) smt))} to a pseudo-boolean instance, the execution path is first to try \textbf{(then simplify sat)}, a linear strategy, for 4 seconds and then to execute \textbf{smt}, another linear strategy. Thus, when evaluating $S_c$ for each input instance $f$, we first convert $S_c$ to its equivalent linear strategy sequence $[S_0, S_1, ..., S_N]$, where the performance of each linear strategy $S_0$, $S_1$, ..., $S_N$ on $f$ is known and cached in the first stage. Then, the performance of $S_c$ can be derived directly from these cached results without further call to the SMT solver. This approach enables MCTS to traverse a significantly larger search space in the second stage, facilitating the discovery of effective complex strategies.  

Another advantage of our staged search method is its adaptiveness to long timeouts. Strategy synthesis typically operates under short solver-instance timeouts, for example, 10 seconds in the \texttt{FastSMT} study. 
Increasing the timeout linearly raises the total synthesis time, making it prohibitively costly to synthesize strategies with extended timeouts, like 5 minutes or more, through direct evaluation.
Our staged search method provides a practical approach to synthesizing strategies under such extended timeouts. In the first stage, the linear strategy candidates are still selected based on their performance with a short timeout period (e.g., 10 seconds), ensuring broad exploration.
These strategies are then re-evaluated under the specified long timeout (e.g., 5 minutes). The second-stage MCTS leverages these re-evaluation results, enabling the synthesis of customized strategies optimized for these extended timeout scenarios.


\section{Experiment Design, Results, and Analysis}
\label{sec:exp}



\subsection{Experimental Design}
We evaluated \texttt{Z3alpha} across six logics, namely QF\_\{BV, NIA, NRA, LIA, LRA, S\}. We benchmarked our performance against the SOTA strategy synthesis tool, \texttt{FastSMT}, as well as the default handcrafted strategy in \texttt{Z3}. \texttt{CVC5} was also included as a baseline for comparison. To assess the robustness of \texttt{Z3alpha}, we designed a series of experiments tailored to different scenarios. 

\noindent{\bf Experimental Design for Specific Classes of Benchmarks:} In Section \ref{sec:exp_indivdual}, we describe the evaluation of the solvers on five important SMT-LIB benchmark sets, namely, \texttt{Sage2} (QF\_BV), \texttt{core} (QF\_BV),  \texttt{AProVE} (QF\_NIA), \texttt{leipzig} (QF\_NIA), and \texttt{hycomp} (QF\_NRA), covering three different logics. These specific benchmark sets were chosen as they were also utilized in the \texttt{FastSMT} study, following the same training-testing split. The size of these benchmark sets varies, ranging from 167 to 7436 instances. A 10-second timeout was chosen for evaluation, a common practice in the SMT strategy synthesis research \cite{stratevo_2016,fastsmt_2018}. This timeout period was also justified as \texttt{Z3alpha} was able to solve over 90\% of the testing instances in four out of the five benchmark sets within this time frame. Additionally, in this section, we included a version of our tool, \texttt{Z3alpha0}, that did not employ staged search, for ablation study purposes.

\noindent{\bf Experimental Design for Benchmarks from Diverse Applications:} In Section \ref{sec:entire_exp}, we extended our experiments to the entire SMT-LIB QF\_LIA and QF\_LRA benchmarks, which are obtained from diverse applications. 
These experiments were intended to evaluate the versatility and adaptability of \texttt{Z3alpha} in handling diverse problem sets. For both QF\_LIA and QF\_LRA, we randomly selected 750 SMT-LIB instances for our training set, and utilized all remaining instances in the logic (12,476 instances and 1,003 instances for QF\_LIA and QF\_LRA, respectively) as our testing set. 

\noindent{\bf Experimental Design for Long Timeout:} In Section \ref{sec:timeout}, we expanded our experiments to include longer timeouts, specifically 1-minute and 5-minute durations. These evaluations were conducted on the most challenging benchmark set, \texttt{Sage2}, from Section \ref{sec:exp_indivdual}. \texttt{Z3alpha} synthesized new strategies for these scenarios, using the method described in Section \ref{sec:staged}. \texttt{FastSMT} used the identical strategy as in Section \ref{sec:exp_indivdual} for these extended timeout cases, adhering to the approach described in its paper.

\noindent{\bf Experimental Design for User-defined Tactics:} \texttt{Z3} allows users to implement new rewrite rules or solver algorithms as tactics. Section \ref{sec:exp_str} evaluated the capability of \texttt{Z3alpha} in synthesizing strategies that integrate both user-defined and built-in tactics. This experiment targeted the QF\_S logic. In recent years, the Z3 String Constraint Solver team \footnote{\url{https://z3string.github.io/}} implemented new tactics, such as the arrangement-based solver \texttt{Z3str3} \cite{z3str3_2017}, the Length Abstraction Solver (LAS) \cite{z3str4_2021}, and specialized rewrite rules for regular expressions \cite{z3strre_2021} for string constraint problems. These tactics are
in addition to the default built-in sequence solver in \texttt{Z3}. \texttt{Z3str4} \cite{z3str4_2021} combines the aforementioned tactics using a meticulously handcrafted strategy. In this experiment, we leveraged \texttt{Z3alpha} to construct a strategy using the same set of tactics as in \texttt{Z3str4} and then compared their performances. \texttt{Z3alpha} was trained on 750 randomly chosen QF\_S instances from SMT-LIB. The testing was conducted on all the remaining 18,173 instances in the logic. 


\subsection{Experimental Setup}
\label{sec:setup}

For every experiment, there were a training instance set and a testing instance set. \texttt{Z3alpha}, as well as \texttt{FastSMT}, synthesized strategies based on the training set while reporting experiment results on the testing set. The approach aligns with our problem statement, wherein our goal is to synthesize a strategy based on a representative set $P$, and the strategy is expected to generalize well.


In each experiment, \texttt{Z3alpha} first selected 20 linear strategies from 800 first-stage MCTS simulations. Subsequently, during the second stage with 300,000 simulations, \texttt{Z3alpha} searched for the most effective strategy that combined these linear strategy candidates. The final synthesized strategy was the one that yielded the lowest PAR-10 score during the second-stage search. 
To keep a similar time budget, the non-staged-search version \texttt{Z3alpha0} ran MCTS for 1,000 simulations with the full CFG $G$, in search of the best strategy (including branching ones). 

\noindent \textbf{Competing Solvers:}
\texttt{Z3alpha} was implemented in Python 3.10 and was integrated with \texttt{Z3-4.12.2}. \texttt{FastSMT} was also integrated with the same version of \texttt{Z3}. Both tools constructed their strategies using the identical tactic and parameter set offered by \texttt{Z3}, and executed these strategies with \texttt{Z3}. 
See the Appendix for a detailed description of the tactic and parameter candidates for each SMT logic. 
Baseline solvers used in the experiment were  \texttt{Z3-4.12.2} and \texttt{CVC5-1.0.5}. We compared our performance with \texttt{FastSMT} in all experiments other than the experiment described in Section \ref{sec:exp_str}, since \texttt{Z3str4} does not provide Python APIs for the user-defined tactics that are required by \texttt{FastSMT}.

\noindent \textbf{Computational Environments:} 
Both our synthesis and testing were conducted on a high-performance CentOS 7 cluster equipped with Intel E5-2683 v4 (Broadwell) processors running at 2.10 GHz, accompanied by 75 gigabytes of memory. 

\noindent \textbf{Variability:} Both the \texttt{Z3alpha} and \texttt{FastSMT} algorithms make use of randomness, leading to the possibility of synthesizing different strategies in separate runs. To account for this variability, the results in Section \ref{sec:exp_indivdual} were average from 5 runs with different random seeds. Since little variability was found in Section \ref{sec:exp_indivdual} and the much more intense computational nature of the later experiments, we only report results from one run in later sections. 


\noindent \textbf{Metrics:} Consistent with the evaluation criteria used in the SMT-COMP, our performance metric was based on the number of correctly solved instances. For clearer comprehension by our readers, we present these results as a percentage, reflecting the proportion of solved instances out of the total tested.  
Additionally, we include results, such as PAR-2 and PAR-10 scores, in the Appendix for further reference.  



\subsection{Analysis of QF\_BV, QF\_NIA, QF\_NRA Results}
\label{sec:exp_indivdual}

The first part of Table \ref{table:solved} summarizes the results on the five selected benchmark sets, namely \texttt{Sage2}, \texttt{core}, \texttt{AProVE}, \texttt{leipzig}, and \texttt{hycomp}, across the QF\_BV, QF\_NIA, and QF\_NRA logics. Notably, \texttt{Z3alpha} surpassed the default \texttt{Z3} strategy, as well as \texttt{CVC5} in all of these benchmark sets, achieving the leading position in four of the five sets among all tested tools. In the particularly challenging QF\_BV benchmark set \texttt{Sage2}, \texttt{Z3alpha} excelled by solving an impressive 42.7\% more instances than the default strategy did. Furthermore, \texttt{Z3alpha} outperformed \texttt{Z3alpha0} across all benchmarks, underscoring the effectiveness of the staged search. The synthesis time for \texttt{Z3alpha} was on par with, and in most experiments, less than, the synthesis time for \texttt{FastSMT}. For instance, while the strategy synthesis for \texttt{AProVE} took 759.6 minutes, \texttt{Z3alpha} completed the task in 293.1 minutes, in which the stage-1 and stage-2 took 213.7 and 79.4 minutes respectively. One key distinction between the synthesized strategies from \texttt{Z3alpha} and \texttt{FastSMT} was that \texttt{Z3alpha} strategies are more interpretable. For example, the \texttt{FastSMT} strategies for \texttt{AProVE} can have more than a thousand branches, while \texttt{Z3alpha} strategies usually have less than five branches. 

\subsection{Analysis of QF\_LIA and QF\_LRA Results}
\label{sec:entire_exp}
When tested across SMI-LIB benchmarks in the entire logic of QF\_LIA and QF\_LRA, \texttt{Z3alpha} also demonstrated consistent performance, as shown in rows of QF\_LIA and QF\_LRA in Table \ref{table:solved}. \texttt{Z3alpha} solved 2.2\% and 3.7\% more instances than \texttt{Z3} in QF\_LIA and QF\_LRA, respectively. While \texttt{FastSMT} solved 4 more instances than \texttt{Z3alpha} in QF\_LRA, its performance suffered significantly in QF\_LIA, solving 58.2\% fewer instances than the default \texttt{Z3} strategy.

\subsection{Analysis of QF\_BV Results with Long Timeout}
\label{sec:timeout}

The results for experiments of 1-minute-timeout and 5-minute-timeout are shown in Table \ref{table:solved}. In every scenario, \texttt{Z3alpha} continued to maintain superior performance compared to both \texttt{FastSMT} and \texttt{Z3}. The performance advantage over \texttt{FastSMT} slightly increased when the timeouts were extended. However, an important shift was the significantly better performance of \texttt{CVC5} over the \texttt{Z3}-based methods for \texttt{Sage2} under long timeouts. This suggests a promising future research direction of extending the strategy synthesis method across different solvers.

\subsection{Results with User-Defined Tactics for QF\_S}
\label{sec:exp_str}

The test results for \texttt{Z3alpha} with user-defined QF\_S tactics are shown in the QF\_S row of Table \ref{table:solved}. \texttt{Z3alpha} demonstrated superior performance over all baseline solvers. Interestingly, the handcrafted  \texttt{Z3str4} strategy, despite employing the same tactic portfolio as \texttt{Z3alpha}, performed worse than the default \texttt{Z3} strategy. The under-performance could be attributed to two factors: (1) the \texttt{Z3str4} strategy was optimized for logics including both QF\_S and QF\_SLIA, which could be sub-optimal for QF\_S alone; (2) the tuning of the \texttt{Z3str4} strategy was carried out on an earlier version of \texttt{Z3}. These points emphasize the importance and benefits of automated strategy customization, tailored for specific problems and updated base solvers. 


\section{Conclusions}
In this work, we present \texttt{Z3alpha}, a novel MCTS-based method for SMT strategy synthesis. \texttt{Z3alpha} introduces layered and staged search heuristics upon the conventional MCTS framework, enabling a low-cost and effective search within the expansive strategy space. The superiority of \texttt{Z3alpha} was demonstrated by extensive experiments across six SMT logics. In all the experiments, \texttt{Z3alpha} consistently surpassed the default \texttt{Z3} solver and outperformed both \texttt{FastSMT} and \texttt{CVC5} in the majority of cases.  

Our method is currently implemented only upon \texttt{Z3}, because other prominent solvers, like \texttt{CVC5}, to the best of our knowledge, do not offer an interface to group preprocessing and solving steps flexibly. We hope our strong empirical results will encourage a universal user-controllable strategy language in the SMT community. There is substantial potential to further enhance solver performance by applying our method across tactics from different solvers, thereby leveraging their complementary strengths.

\section*{Acknowledgements}
We thank Kate Larson, Arie Gurfinkel, and Mark Crowley for their valuable feedback; Kaihang Jiang for his contribution to code testing and data collection; and the Digital Research Alliance of Canada for providing the computational resources and technical support. 

The work of Zhengyang Lu is supported by the Engineering Excellence Doctoral Fellowship (EEDF) at the University of Waterloo; the work of Florin Manea was supported by the DFG-Heisenberg grant no. 466789228.

\bibliographystyle{named}
\bibliography{ijcai24}

\clearpage

\appendix

\section*{Appendix}


\section{Additional Methodology Details}

\textbf{Domain Knowledge Rules:} In Section \ref{sec:framework}, we mention that, in order to reduce the action space, additional domain-knowledge-based simplification rules are introduced to the problem modeling. Such rules include:

\begin{enumerate}
    \item When $\langle \text{Strategy} \rangle \rightarrow \langle \text{Tactic} \rangle$ is applied, $\langle \text{Tactic} \rangle$ can only choose a solver-wrapper tactic, e.g., \textbf{smt}. When $\langle \text{Strategy} \rangle \rightarrow \textbf{(then} \: \langle \text{Tactic} \rangle \: \langle \text{Strategy} \rangle \textbf{)}$ is applied, solver-wrapper tactics cannot be chosen for $\langle \text{Tactic} \rangle$.

    \item Do not apply the \textbf{try-for} rule for a strategy that has already been set for a \textbf{try-for} timeout.

    \item The \textbf{if} predicate can only appear at the first three depths of the constructed strategy's syntax tree. Moreover, do not apply the \textbf{if} rule after any tactic application, since tactic applications may alter the formula measures, disabling the staged search. 

    \item Certain candidate values are preselected for numerical tactic parameters, \textbf{try-for} timeouts, and numerical probe comparisons.
    
    \item Do not apply the tactic \textbf{nla2bv} more than once in a sequence of tactic applications.

    \item Apply the tactic \textbf{bit-blast} only immediately after applying the tactic \textbf{simplify}.
    
\end{enumerate}

We recognize that certain simplification rules might eliminate parts of the search space that include effective strategies. Nevertheless, these rules are instrumental in substantially reducing the vast search space and enabling efficient search techniques like the stage search. Our empirical experimental results strongly indicate that the remaining space still encompasses robust strategies. 

\noindent\textbf{Tactic and Parameter Pools:} We list the pool of tactic and parameter candidates for each tested SMT logic in a series of tables, specifically from Table \ref{table:tac_qfbv} to Table \ref{table:tac_qfs}. Note that the candidate selection is identical for both \texttt{Z3alpha} and \texttt{FastSMT}. 



\begin{figure}[t]
\centering
\includegraphics[width=1.1\columnwidth]{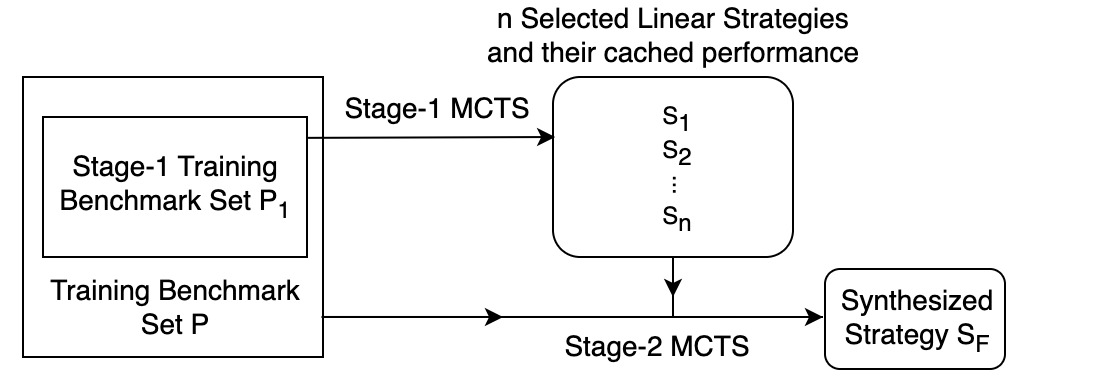}
\caption{The Staged MCTS Framework}
\label{fig:staged}
\end{figure}

\noindent\textbf{Staged Search Framework: } Figure \ref{fig:staged} illustrates the staged MCTS framework. A set of $n$ linear strategies is selected from the first-stage MCTS, and these linear strategies are synthesized into one final combined strategy $S_F$ in the second-stage MCTS. Note that we may opt for a subset $P_1$ of the complete training benchmark set $P$ for the first stage. This is because each stage-1 MCTS simulation can be expensive for a large training set, since it involves calling an SMT solver for every training instance. After selecting linear strategies using a smaller subset, we can evaluate the selected strategies across all instances in $P$, with results cached for the second stage. This approach eliminates the resource-intensive need to evaluate every strategy explored in the first stage on the full set $P$, while still providing larger-scale evaluation data for the second-stage MCTS.

\begin{table*}[t]
\centering
\fontsize{9pt}{11pt}\selectfont
\setlength{\tabcolsep}{5pt}
\begin{tabular}{cllccccccc}
\hline
\textbf{Timeout(s)} & \multicolumn{1}{c}{\textbf{Logic}} & \multicolumn{1}{c}{\textbf{Benchmark}}  & \textbf{Test Size} & \textbf{\texttt{Z3alpha}} & \textbf{\texttt{Z3alpha0}} & \textbf{\texttt{FastSMT}}  & \textbf{\texttt{Z3}} & \textbf{\texttt{CVC5}}& \textbf{\texttt{Z3str4}}\\ \hline
\multirow{8}{*}{10}  & \multirow{2}{*}{QF\_BV}   & \texttt{Sage2} & 6444& \textbf{11.01} & 12.97 & \underline{11.30}     & 13.68     & 14.02 & -  \\ 
  &   & \texttt{core} &270 & \underline{1.06}&1.14& \textbf{0.89}&  6.08& 5.21 & -\\ \cline{2-10} 
& \multirow{2}{*}{QF\_NIA}& \texttt{AProVE} & 1712& \textbf{1.27} &\underline{1.64}& 2.89  &  3.28 & 6.62 &-  \\ 
   &  & \texttt{leipzig} & 68 & \textbf{2.13}  & \underline{2.58} & 2.87 & 2.63 & 15.72 &-\\ \cline{2-10} 
 & QF\_NRA & \texttt{hycomp} & 1982& \textbf{1.90}  & \underline{1.95}  & 2.29 & 3.19 & 3.17 &-  \\ \cline{2-10} 
 & QF\_LIA & entire logic &12476 & \textbf{5.23}  & -  & 13.88  & \underline{5.52} & 7.61  &- \\ \cline{2-10} 
 & QF\_LRA  & entire logic&1003 & \underline{5.93}  & -  & \textbf{5.92}    & 6.46    & 7.74  &- \\ \cline{2-10} 
 & QF\_S & entire logic&18173& \textbf{0.41}  & -  & -  & 0.57 & \underline{0.42} & 0.71  \\ \hline
60    & QF\_BV  & \texttt{Sage2} & 6444& \underline{43.57}  & - & 43.82 & 57.41 & \textbf{38.77} &-\\ \hline
300   & QF\_BV   & \texttt{Sage2} & 6444& \underline{163.26}  & - & 181.00  &  214.23 & \textbf{80.71} &-       
\\ \hline              
\end{tabular}
\caption{{\bf \texttt{Z3alpha} vs. SOTA Solvers:} Average PAR-2 score on the selected SMT-LIB benchmarks across six SMT logics (
In each experiment, the result of the leading tool is highlighted in bold, while the second-best is underscored.)
}
\label{table:par2}
\end{table*}

\begin{table*}[t]
\centering
\fontsize{9pt}{11pt}\selectfont
\setlength{\tabcolsep}{5pt}
\begin{tabular}{cllccccccc}
\hline
\textbf{Timeout(s)} & \multicolumn{1}{c}{\textbf{Logic}} & \multicolumn{1}{c}{\textbf{Benchmark}}  & \textbf{Test Size} & \textbf{\texttt{Z3alpha}} & \textbf{\texttt{Z3alpha0}} & \textbf{\texttt{FastSMT}}  & \textbf{\texttt{Z3}} & \textbf{\texttt{CVC5}}& \textbf{\texttt{Z3str4}}\\ \hline
\multirow{8}{*}{10}  & \multirow{2}{*}{QF\_BV}   & \texttt{Sage2} & 6444& \textbf{48.66} & 59.97 & \underline{50.27}     & 64.01     & 64.71 & -  \\ 
  &   & \texttt{core} &270 & \underline{1.36}& 1.49& \textbf{0.89}& 25.64& 19.14 & -\\ \cline{2-10} 
& \multirow{2}{*}{QF\_NIA}& \texttt{AProVE} & 1712& \textbf{5.45} &\underline{7.64}& 10.63  &  11.27 & 30.69 &-  \\ 
   &  & \texttt{leipzig} & 68 & \textbf{9.19}  & 11.29 & 12.28 & \underline{10.87} & 75.72 &-\\ \cline{2-10} 
 & QF\_NRA & \texttt{hycomp} & 1982& \textbf{8.82}  & \underline{9.27}  & 10.98 & 14.69 & 14.75 &-  \\ \cline{2-10} 
 & QF\_LIA & entire logic &12476 & \textbf{24.23}  & -  & 68.93 & \underline{25.81} & 35.78  &- \\ \cline{2-10} 
 & QF\_LRA  & entire logic&1003 & \underline{27.06}  & -  & \textbf{26.74}    & 29.67    & 36.85  &- \\ \cline{2-10} 
 & QF\_S & entire logic&18173& \textbf{1.23}  & -  & -  & 2.00 & \underline{1.72} & 2.96 \\ \hline
60    & QF\_BV  & \texttt{Sage2} & 6444& \underline{188.30}  & - & 198.68 & 260.47 & \textbf{145.52} &-\\ \hline
300   & QF\_BV   & \texttt{Sage2} & 6444& \underline{743.14}  & - & 837.61  & 940.12 & \textbf{240.86} &-       
\\ \hline              
\end{tabular}
\caption{{\bf \texttt{Z3alpha} vs. SOTA Solvers:} Average PAR-10 score on the selected SMT-LIB benchmarks across six SMT logics (
In each experiment, the result of the leading tool is highlighted in bold, while the second-best is underscored.)
}
\label{table:par10}
\end{table*}

\noindent\textbf{Selection Criteria of Linear Strategies: } After the first-stage MCTS, we select $n$ linear strategies for the second-stage synthesis. This selection is performed iteratively from the pool of linear strategies explored in the first stage using a greedy algorithm. The idea is to incrementally build a linear strategy portfolio that collectively performs the best. Here, we define the {\it virtual best strategy} for a strategy set as an oracle that, for each instance, perfectly selects the best strategy from the strategy set without any overhead. Then, the selection criterion is to select the linear strategy that, when added to the incumbent selected strategy set, maximizes the {\it virtual best strategy} performance for a given instance set. Starting from an empty set, we apply this criterion to add one linear strategy at a time to the set until $n$ strategies are selected. Consistent with the entire synthesis process, performance is measured by the average PAR-10 score.

\section{Additional Experimental Details}
\subsection{Training and Testing Benchmarks} 
In experiments for specific benchmark classes (Section \ref{sec:exp_indivdual}), we used the same experimental benchmark sets that were used in the \texttt{FastSMT} study \cite{fastsmt_2018}. They divided each benchmark set, i.e., \texttt{Sage2}, \texttt{core},  \texttt{AProVE}, \texttt{leipzig}, and \texttt{hycomp}, into a training set, a validation set, and a test set. For \texttt{Z3alpha} synthesis, we retained the test set for evaluation and merged the training and validation sets to form the \texttt{Z3alpha} training benchmark set $P$. Within it, the \texttt{FastSMT} training set was specifically used as the training subset $P_1$ for the first-stage MCTS. 

For QF\_LIA, QF\_LRA, and QF\_S experiments, we randomly selected 750 instances to constitute the \texttt{Z3alpha} training set $P$, while designating all remaining instances in the SMT-LIB as the test set. From $P$, 250 instances were chosen as the stage-1 training subset $P_1$. Our experiments showed that this training size was sufficient to develop robust strategies for the logic. In these experiments, $P_1$ served as the training set for \texttt{FastSMT}, and the set difference $P \setminus P_1$ constituted the validation set, aligning with the \texttt{FastSMT} benchmark split sizes for large sets. All competing solvers were evaluated on the same test set.



\subsection{Experimental Results in PAR-2 and PAR-10}
Table \ref{table:par2} and Table \ref{table:par10} show the experimental results as average PAR-2 and PAR-10 scores, respectively. The PAR-2 score was calculated based on the solver runtime for each successfully solved instance in the test set, and a penalty of twice the timeout for unsolved instances. Similarly, the PAR-10 score imposed a penalty of ten times the timeout for the unsolved. A lower score in both PAR-2 and PAR-10 indicates superior performance. Notably, the PAR-2 and PAR-10 results strongly aligned with the results of the number of solved instances, as presented in Table \ref{table:solved}.


\begin{table}[t]
\centering
\begin{tabular}{l c}
\hline
\multicolumn{1}{c}{\textbf{Tactic}} & \multicolumn{1}{c}{\textbf{Parameter}}\\
\hline\hline
\multirow{8}{*}{simplify} &  elim\_and \\ 
& blast\_distinct \\
& local\_ctx \\
& som \\
& flat \\
& pull\_cheap\_ite \\
& hoist\_mul\\
& push\_ite\_bv \\
\hline
propagate-values & push\_ite\_bv \\
\hline
ctx-simplify & - \\
\hline
elim-uncnstr & - \\
\hline
solve-eqs & - \\
\hline
purify-arith & - \\
\hline
max-bv-sharing & - \\
\hline
aig & - \\
\hline
reduce-bv-size & - \\
\hline
ackermannize\_bv & - \\
\hline
bit-blast & - \\
\hline
smt & random\_seed \\
\hline
sat & - \\
\hline
qfbv & - \\
\hline
\end{tabular}
\caption{Selected tactic and parameter candidates for QF\_BV}
\label{table:tac_qfbv}
\end{table}

\begin{table}[t]
\centering
\begin{tabular}{l c}
\hline
\multicolumn{1}{c}{\textbf{Tactic}} & \multicolumn{1}{c}{\textbf{Parameter}}\\
\hline\hline
\multirow{9}{*}{simplify} &  elim\_and \\ 
& blast\_distinct \\
& local\_ctx \\
& som \\
& flat \\
& hi\_div0 \\
& pull\_cheap\_ite \\
& hoist\_mul\\
& push\_ite\_bv \\
\hline
propagate-values & push\_ite\_bv \\
\hline
ctx-simplify & - \\
\hline
elim-uncnstr & - \\
\hline
solve-eqs & - \\
\hline
max-bv-sharing & - \\
\hline
nla2bv & nla2bv\_max\_bv\_size \\
\hline
bit-blast & - \\
\hline
smt & random\_seed \\
\hline
\multirow{3}{*}{qfnra-nlsat} & inline\_vars \\
& factor \\
& seed \\
\hline
lia2card & - \\
\hline
card2bv & - \\
\hline
cofactor-term-ite & - \\
\hline
qfnia & - \\
\hline
\end{tabular}
\caption{Selected tactic and parameter candidates for QF\_NIA}
\label{table:tac_qfnia}
\end{table}

\begin{table}[t]
\centering
\begin{tabular}{l c}
\hline
\multicolumn{1}{c}{\textbf{Tactic}} & \multicolumn{1}{c}{\textbf{Parameter}}\\
\hline\hline
\multirow{9}{*}{simplify} &  elim\_and \\ 
& blast\_distinct \\
& local\_ctx \\
& som \\
& flat \\
& hi\_div0 \\
& pull\_cheap\_ite \\
& hoist\_mul\\
& push\_ite\_bv \\
\hline
propagate-values & push\_ite\_bv \\
\hline
ctx-simplify & - \\
\hline
elim-uncnstr & - \\
\hline
solve-eqs & - \\
\hline
max-bv-sharing & - \\
\hline
nla2bv & nla2bv\_max\_bv\_size \\
\hline
bit-blast & - \\
\hline
smt & random\_seed \\
\hline
\multirow{3}{*}{qfnra-nlsat} & inline\_vars \\
& factor \\
& seed\\
\hline
qfnra & - \\
\hline
\end{tabular}
\caption{Selected tactic and parameter candidates for QF\_NRA}
\label{table:tac_qfnra}
\end{table}

\begin{table}[t]
\centering
\begin{tabular}{l c}
\hline
\multicolumn{1}{c}{\textbf{Tactic}} & \multicolumn{1}{c}{\textbf{Parameter}}\\
\hline\hline
\multirow{9}{*}{simplify} &  elim\_and \\ 
& blast\_distinct \\
& local\_ctx \\
& som \\
& flat \\
& pull\_cheap\_ite \\
& push\_ite\_arith \\
& hoist\_ite \\
& arith\_lhs \\
\hline
propagate-values & push\_ite\_bv \\
\hline
ctx-simplify & - \\
\hline
elim-uncnstr & - \\
\hline
solve-eqs & - \\
\hline
propagate-ineqs & - \\
\hline
\multirow{2}{*}{add-bounds} & add\_bound\_lower \\
& add\_bound\_upper \\
\hline
normalize-bounds & - \\
\hline
lia2pb & lia2pb\_max\_bits \\
\hline
smt & random\_seed \\
\hline
qflia & - \\
\hline
\end{tabular}
\caption{Selected tactic and parameter candidates for QF\_LIA}
\label{table:tac_qflia}
\end{table}

\begin{table}[t]
\centering
\begin{tabular}{l c}
\hline
\multicolumn{1}{c}{\textbf{Tactic}} & \multicolumn{1}{c}{\textbf{Parameter}}\\
\hline\hline
\multirow{5}{*}{simplify} &  elim\_and \\ 
& blast\_distinct \\
& local\_ctx \\
& som \\
& flat \\
\hline
propagate-values & - \\
\hline
ctx-simplify & - \\
\hline
elim-uncnstr & - \\
\hline
solve-eqs & - \\
\hline
smt & random\_seed \\
\hline
qflra & - \\
\hline
\end{tabular}
\caption{Selected tactic and parameter candidates for QF\_LRA}
\label{table:tac_qflra}
\end{table}

\begin{table}[t]
\centering
\begin{tabular}{l c}
\hline
\multicolumn{1}{c}{\textbf{Tactic}} & \multicolumn{1}{c}{\textbf{Parameter}}\\
\hline\hline
\multirow{5}{*}{simplify} &  elim\_and \\ 
& blast\_distinct \\
& local\_ctx \\
& som \\
& flat \\
\hline
propagate-values & - \\
\hline
ctx-simplify & - \\
\hline
elim-uncnstr & - \\
\hline
solve-eqs & - \\
\hline
ext\_str & - \\
\hline
ext\_strSimplify & - \\
\hline
ext\_strToRegex & - \\
\hline
ext\_strToWE & - \\
\hline
arr & - \\
\hline
las & - \\
\hline
smt & random\_seed \\
\hline
\end{tabular}
\caption{Selected tactic and parameter candidates for QF\_S}
\label{table:tac_qfs}
\end{table}


\end{document}